\pdfoutput=1
\documentclass[11pt]{article}

\usepackage{acl}

\usepackage{times}
\usepackage{latexsym}

\usepackage[T1]{fontenc}
\usepackage{mathtools}
\usepackage{bbm}
\usepackage{amssymb}
\usepackage[super]{nth}

\usepackage[utf8]{inputenc}
\usepackage{pgfplots}
\usepackage{adjustbox}
\usepackage{multicol}
\usepackage{multirow}
\usepackage{graphicx}
\usepackage{cleveref}
\usepackage{booktabs}
\usepackage{colortbl}
\usepackage{tabulary}
\usepackage{etoolbox}

\usepackage{microtype}

\usepackage{inconsolata}

\title{Accurate Knowledge Distillation with \emph{n}-best Reranking}

\author{Hendra Setiawan \\
  Apple \\
  \texttt{hendra@apple.com}}

\begin{document}
\maketitle
\begin{abstract}

We propose utilizing \emph{n}-best reranking to enhance Sequence-Level Knowledge Distillation \cite{kim-rush-2016-sequence} where we extract pseudo-labels for student model's training data from top \emph{n}-best hypotheses and leverage a diverse set of models with different inductive biases, objective functions or architectures, including some publicly-available large language models, to pick the highest-quality hypotheses as labels. 
The effectiveness of our proposal is validated through experiments on the WMT'21 German $\leftrightarrow$ English and Chinese $\leftrightarrow$ English translation tasks. 
Our results demonstrate that utilizing pseudo-labels generated by our \emph{n}-best reranker leads to a significantly more accurate student model. 
In fact, our best student model achieves comparable accuracy to a large translation model from \cite{tran-etal-2021-facebook} with 4.7 billion parameters, while having two orders of magnitude fewer parameters.

\end{abstract}

\section{Introduction}

Knowledge Distillation (KD) \cite{hinton-distilling,modelcompression} plays a pivotal role in many machine learning tasks including neural machine translation (NMT). 
This is evident in recent translation evaluations \cite{akhbardeh-etal-2021-findings,kocmi-etal-2022-findings,agrawal-etal-2023-findings}, where the majority of submissions incorporate KD into their training pipelines.
Central to this paper is KD's primary strength, which lies in its ability to utilize a larger \emph{teacher} model to train a smaller \emph{student} model effectively. 
Consequently, the accuracy of the teacher model correlates with that of the student model.

One well-known yet simple approach for enhancing the accuracy of machine learning models involves model ensembling \cite{dietterich2000ensembles}, as also has been applied by \cite{hinton-distilling} for KD. 
In this approach, the underlying models are typically trained on the same datasets but with varying random initializations.
Many existing works applying KD to NMT have adopted this approach, including its dominant variant, namely sequence-level KD \cite{kim-rush-2016-sequence}, which deploys an ensemble of teacher models to generate the \emph{pseudo-labels} of the student models' training data.
To maintain the simplicity of the inference process when generating the labels, the sequence-level KD often imposes the requirements that the underlying models share the same vocabulary and network architecture.
These factors, unfortunately, restrict the types of models that can be included in the ensemble, thereby limiting the avenues for improving the teacher accuracy. 

We introduce an \emph{n}-best reranking approach to extend the vanilla sequence-level KD to allow KD to benefit from a more diverse type of models.
\emph{N}-best reranking is a well-known and powerful approach for boosting translation accuracy, as evident in \cite{marie-etal-2020-combination,qian-etal-2021-volctrans,tran-etal-2021-facebook} just to name a few.
Unfortunately, the associated inference cost can be too high for this approach to be deployed in online environments with tight latency constraint. 
By applying \emph{n}-best reranking for distillation, the elevated computational cost is shifted to training stage without affecting the latency of the deployed student model, thus sharing a similar motivation as \cite{yang-etal-2022-nearest,finkelstein2023mbr}.

Our proposed approach involves a two-step process. 
The first step involves generating a high-quality \emph{n}-best list from each source sentence. 
Our initial study indicates a potential gain of almost 10 BLEU on our validation set if we consider hypotheses beyond top-1.
The final decision on which hypothesis to be selected is deferred to the second step where  we take advantage of models with various architectures, inductive biases, sources of training data and objective functions to rerank the \emph{n}-best list.
To increase model diversity further, we also incorporate open-source large pretrained models in our experiments, tapping into their availability and increased translation capabilities.
This flexibility stands in contrast to the rigid type of teacher models deployed by the vanilla sequence-level KD.

We showcase our reranker's effectiveness in two scenarios, namely the traditional and iterative KDs. 
In the former, pseudo-labels are directly utilized for training student models, while the latter, often dubbed as \emph{self-training}, involves an extra step of iteratively retraining the teacher models using the pseudo-labels \cite{li-etal-2019-niutrans}. 
In the first scenario, we thoroughly examine the accuracy of student models trained with pseudo-labels generated by our reranker, comparing them to models trained with pseudo-labels from the vanilla sequence-level KD. 
In the second scenario, we investigate whether our reranker can yield a valuable cascading effect by progressively enhancing teacher models using the pseudo labels from teacher models from the previous iteration. 
These improved teacher models, in turn, may generate superior pseudo-labels for training more accurate student models.

We also explore efforts that are directed at scaling up our method to distill large-scale training data more efficiently. 
This includes methods such as \emph{model selection}, where we choose a smaller set of models for distillation that results in only a minimal accuracy drop, and \emph{transfer set reduction}, where we decrease the volume of data to be distilled to only include in-domain samples.
More concretely, we conduct extensive experiments on WMT'21 German $\leftrightarrow$ English and Chinese $\leftrightarrow$ English translation tasks. 
Our final model is as accurate as a large multilingual model with 4.7 billion parameters, despite having only 68 million parameters.

\section{Background: Sequence-Level KD}

KD trains a student model ($p_\theta$) with the supervision of a more powerful teacher model ($q_{\theta}$) by minimizing the discrepancy between the prediction of the student model with that of the teacher model. 
%To train a student model ($p_\theta$) with the supervision of a teacher model ($q_{\theta}$), KD seeks to minimize the discrepancy between the prediction of the student model with that of the teacher model. 
Sequence-level KD, proposed by \newcite{kim-rush-2016-sequence}, extends KD by minimizing the discrepancy at the level of \emph{sequence} (rather than at token level) by introducing the following loss function:
\begin{align*}
\mathcal{L}_{\text{SEQ-KD}} = - \sum_{t \in \mathcal{T}} q_{\theta}(\boldsymbol{t}|\boldsymbol{s}) \log p_{\theta}(\boldsymbol{t}|\boldsymbol{s})
\end{align*}
\noindent where $\boldsymbol{t} \in \mathcal{T}$ represents the set of all possible sequences that the teacher model can generate from a source sentence ($\boldsymbol{s}$). 

Since enumerating all possible sequences is intractable, \newcite{kim-etal-2021-distilling-knowledge} approximate the distribution with its mode $\hat{\boldsymbol{t}}$ and arrive at the following approximation:
\begin{align*} 
 \mathcal{L}_{\text{SEQ-KD}} & \approx - \mathbbm{1}\left\{\boldsymbol{t} = \hat{\boldsymbol{t}}\right\} \log p_{\theta}%(\boldsymbol{t}|\boldsymbol{s}) \\
 & \approx - \log p_{\theta}(\hat{\boldsymbol{t}}|\boldsymbol{s}),
\end{align*}
where the mode is obtained via the following inference: $\hat{\boldsymbol{t}}=\arg\max q_{\theta}(\boldsymbol{t}|\boldsymbol{s})$. 
This simple approximation allows NMT to reuse the same standard training pipeline for student model with only a slight modification, namely by substituting the original labels $\boldsymbol{t}$ with pseudo-labels $\hat{\boldsymbol{t}}$ when computing the loss function.

\section{\emph{N}-best Reranking for Distillation}

Our \emph{n}-best reranker formulates $q_{\theta}(\boldsymbol{t}|\boldsymbol{s})$ as a weighted log-linear model, which is parameterized by a collection of $M$ models, $\boldsymbol{\mathcal{M}}(\boldsymbol{s}, \boldsymbol{t}) \in \mathbb{R}^M$,  and their associated weights $\boldsymbol{\lambda} \in \mathbb{R}^M$.
Each model, $\mathcal{M}_i(\boldsymbol{s}, \boldsymbol{t}) \in \mathbb{R}$, assigns a real-valued score that indicates the plausibility of a hypothesis $\boldsymbol{t}$ being the translation of $\boldsymbol{s}$ according to the model, hence we refer to $\mathcal{M}(\boldsymbol{s}, \boldsymbol{t})$ as scoring models.
The score of each scoring model $\mathcal{M}_i(\boldsymbol{s}, \boldsymbol{t})$ is weighted by $\lambda_i$ and then combined with all other models to produce the final score.
While the models are considered static, their associated weights are \emph{trainable} parameters, learned via discriminative training. 
We discuss the process in \Cref{sec:optimization}.

To generate pseudo-labels, our reranker applies the following $\arg\max$ formula:
\begin{align}
    \hat{\boldsymbol{t}} = \arg\max_{\boldsymbol{t} \in \mathcal{N}(\boldsymbol{s})} \boldsymbol{\lambda} \cdot \log \boldsymbol{\mathcal{M}}(\boldsymbol{s}, \boldsymbol{t})^\intercal, \label{eq:loglinear} 
\end{align}
\noindent where $\boldsymbol{t} \in \mathcal{N}(\boldsymbol{s})$ refers a hypothesis in an \emph{n}-best list. 
We generate the \emph{n}-best list by running beam search inference with a beam size set to $n$; however, it can also be generated using alternative methods such as epsilon sampling.
We refer to the models used to generate $\mathcal{N}$ as $\boldsymbol{\mathcal{G}}(\boldsymbol{s}) \subset \boldsymbol{\mathcal{M}}(\boldsymbol{s}, \boldsymbol{t})$. %, which is also used to score the \emph{n}-best list. 
If $\boldsymbol{\mathcal{G}}(\boldsymbol{s}) = \boldsymbol{\mathcal{M}}(\boldsymbol{s}, \boldsymbol{t})$ and consists of only one identical translation model, then Eq~\ref{eq:loglinear} would revert back to the vanilla sequence-level KD. 

\subsection{Discriminative Training of $\boldsymbol{\lambda}$}
\label{sec:optimization}

To find the optimal $\boldsymbol{\lambda}$, we utilize discriminative training and a \emph{tuning set}, which is assumed to be drawn from the same distribution of the test sets. 
For this paper, we employ the Margin Infused Relaxed Algorithm (MIRA) \cite{chiang-etal-2008-online}, known for its wide adaptation in Statistical Machine Translation and its ability to handle tens of thousands of inputs.
Without loss of generality, we use BLEU \cite{papineni-etal-2002-bleu} to measure translation accuracy in our experiments. 

MIRA seeks to find $\boldsymbol{\lambda}$ that minimizes the following structured hinge loss $\mathcal{L}_{\text{MIRA}}(\boldsymbol{\lambda})$
\begin{equation}
     = \max_{\boldsymbol{t} \in \mathcal{N}} \Bigl[ \Delta(\boldsymbol{t}) + \boldsymbol{\lambda}\cdot \left( \boldsymbol{\mathcal{M}}(\boldsymbol{s}, \boldsymbol{t})^\intercal - \boldsymbol{\mathcal{M}}(\boldsymbol{s}, \boldsymbol{t^*})^\intercal \right) \Bigr]
     \nonumber
\end{equation}
\noindent where $\boldsymbol{t^*}$ is the \emph{oracle} hypothesis, which refers the hypothesis in the \emph{n}-best list that attains the highest BLEU score, while $\Delta(\boldsymbol{t})$ signifies the BLEU differentials of a hypothesis $\boldsymbol{t}$ with the aforementioned oracle hypothesis. 
Ideally, the optimal $\boldsymbol{\lambda}$ is achieved when the loss reaches 0, indicating that a clear separation can be established between each non-oracle hypothesis and the oracle hypothesis with a margin proportional to their respective BLEU differentials.

In our experiments, we use a variant of MIRA with an efficient batch-level support, called KB-MIRA \cite{cherry-foster-2012-batch} which can be found in the Moses toolkit  \cite{moses}. 
It also includes a sparsity-inducing regularization which we utilize for model selection. 
For a more in-depth discussion on MIRA and its variants, we refer the readers to the cited papers.

\subsection{$\boldsymbol{\mathcal{M}}(\boldsymbol{s}, \boldsymbol{t})$: Models for Scoring \emph{n}-best List}
\label{sec:model}
\renewcommand{\arraystretch}{1.25}
\begin{table*}[ht]
    \centering
    %\medium
    \begin{tabular}{lll}
    \toprule
         Model category & Formulation & Description\\
         \midrule
         (1) Forward translation model (TM) & $\sum_{i=0}^I \log p(t_i|t_{<i},\boldsymbol{s})$ & Variants: Deep Encoder, MEGA \\
         (2) Backward TM & $\sum_{j=0}^J \log p(s_j|s_{<j},\boldsymbol{t})$ & \emph{idem} \\
         (3) Right-to-left model & $\sum_{i=0}^I \log p(t_i|t_{>i},\boldsymbol{s})$ & Applied to (1 \& 2) as well \\
         (4) Domain-adapted TM & $\sum_{i=0}^I \log p(t_i|t_{<i},\boldsymbol{s},d)$ & $d$: domain tag, applied to TM\\
         \hline
         (5) Monolingual language model & $\sum_{i=0}^N \log p(t_i|t_{<i})$ & Causal language model \\
         (6) Alignment-based model & $\log P(\boldsymbol{t}|\boldsymbol{s},a)$ & $a$: token alignment between $y,s$ \\
         (7) MBR loss function & $\mathcal{U}(\boldsymbol{t}|\boldsymbol{t}^\prime\in \mathcal{N})$ & $\mathcal{U} \in$ BLEU, TER, chrF, etc\\
         \multicolumn{3}{l}{(8) Various publicly-available pretrained models, e.g. LASER, mBART, M2M, BLOOM-Z, etc }\\ 
         \bottomrule
    \end{tabular}
    \caption{Categories of models to evaluate hypothesis pair $\boldsymbol{s}, \boldsymbol{t} \in \mathcal{N}$ in the \emph{n}-best list. The first four categories correspond to \emph{in-house} translation models, while the last four correspond to general models.}
    \label{tab:models_cat}
\end{table*}
\renewcommand{\arraystretch}{1}

The efficacy of our \emph{n}-best reranker depends on the diversity and quality of the deployed scoring models.
The log-linear formulation in \Cref{eq:loglinear} imposes minimal assumptions about the underlying models, which relaxes the requirements for the models to strictly adhere to probabilistic principles or comprehensively describe the entire translation process. 
As a result, our reranker is able to accommodate a wide spectrum of models, including heuristics or target-side language models, as long as they assign a relatively meaningful score. 

In total, we conduct experiments involving over 50 scoring models for each language pair.
For conciseness, we group the models into 8 categories and provide a description for each category, summarized in Table~\ref{tab:models_cat}. 
The first four categories include a diverse range of \emph{in-house} translation models, characterized by distinctions in translation directions, generation orders, network architectures, and domain adaptability.
It is worth mentioning that most of these models are developed for other exploratory projects and including them as scoring models provides another avenue to reuse them.
Meanwhile, the last four categories encompass models that do not strictly pertain to translation models but capture some specific nuances of translation phenomena, such as the fluency of hypotheses or the level of agreement between hypotheses. 
%This versatility is facilitated by the log-linear formulation of our \emph{n}-best reranker.

The first category is the \emph{forward translation model} (TM), which includes standard NMT models $p(\boldsymbol{s}|\boldsymbol{t})$. 
These models correspond to an autoregressive translation process that generates the translation sequentially one token at a time conditioned on the source sentence and previously generated tokens. The models in this category include NMT models with various well-known architectures, such as Transformer Big \cite{vaswani-transformer}, Deep Encoder Shallow Decoder \cite{kong-etal-2021-multilingual}, Nearest-Neighbor \cite{khandelwal2021nearest} and MEGA \cite{ma2023mega}. 

The second category is the \emph{backward TM}, which includes models that are similar to the forward TM but with different translation direction. 
In particular, these models focus on modeling the backward translation direction $p(\boldsymbol{t}|\boldsymbol{s})$, useful to capture how likely the source sentences be the translation of a hypothesis, complementing the forward TM.

The third category is the \emph{right-to-left} TM, which includes models from the first two categories but look at a \emph{reverse} generation order, namely generating tokens in a right-to-left fashion. 
According to \cite{liu-etal-2016-agreement,zhou-etal-2019-synchronous}, the left-to-right models are more effective at generating accurate prefixes while the right-to-left models are more effective at generating accurate suffixes. 

The fourth category is \emph{domain-adapted} models, which consists of translation models from the previous three categoeis that we adapt to multiple domains. 
In our experiments, we simply equate the corpus provenance as the domain. 
We adopt a tag-based approach and prepend the source sequence with $d \in \left\{\text{europarl}, \text{commoncrawl}, \text{rapid}, \cdots \right\}$, like in \cite{johnson-etal-2017-googles,ha-etal-2017-effective}.  

The fifth category is the \emph{language model}, consisting of the models that focus on evaluating the fluency aspect of the hypotheses. In our experiments, we train a causal language model with the GPT-2 architecture \cite{radford2019language} on the target side of our parallel data and the monolingual data. Meanwhile, the sixth category is the \emph{alignment} models, consisting of the models that evaluate the fine-grained correspondences between tokens in the hypothesis and the source sequence. To generate the alignment, we use the IBM model 3 \cite{brown-etal-1993-mathematics} from the \texttt{eflomal} toolkit \cite{Ostling2016efmaral}. The seventh category corresponds to the \emph{Minimum Bayes-Risk (MBR) utility function}. Via the models in this category, our reranker is able to give preferences to hypotheses that have the higher level of consensus with other hypotheses in the \emph{n}-best list, as measured by some extrinsic translation metrics. These models infuse our reranker with elements of consensus decoding \cite{kumar-byrne-2004-minimum}.

Our last category consists of various publicly-available pretrained models. 
It includes the LASER sentence-embedding model \cite{DBLP:journals/corr/abs-1812-10464}, the mBART multilingual translation model \cite{liu2020multilingual}, the M2M-100 \cite{fan2020beyond} and the NLLB \cite{nllb2022}. 
It also includes a single dense multilingual model from the WMT21 winning team, namely Facebook AI Research (FAIR) WMT21 \cite{tran-etal-2021-facebook} - currently known as Meta AI Research. 
Additionally, it includes multilingual large language models from BigScience, namely BLOOMZ and mT0 \cite{muennighoff2022crosslingual}. 
These models are trained with significantly more data and not all of them are explicitly trained to optimize translation objectives. 
When utilizing these models, we condition them for translation by prepending five translation examples as the prefix (5-shot) like in \cite{moslem-etal-2023-adaptive}.
The sizes of these models vary from 50 million to 10+ billion parameters, which is larger than the models in other categories. 

\subsection{$\boldsymbol{\mathcal{G}}(\boldsymbol{s})$: Models for Generating \emph{n}-best List }

The efficacy of our \emph{n}-best reranking also hinges upon the accuracy and diversity of the \emph{n}-best list. 
While an ideal scenario involves deploying all scoring models within $\boldsymbol{\mathcal{M}}(\boldsymbol{s},\boldsymbol{t})$, this proves to be both computationally intensive and impractical, especially considering that not all models explicitly generate translations, such as language models. 

To strike a balance between efficiency and efficacy, we choose to utilize the two specific models, which we call the \emph{L2R} and the \emph{R2L} models. 
The L2R model comprises an ensemble of four Transformer Big models \cite{vaswani-transformer} while the R2L model is its right-to-left counterpart. 
The former belongs to the first category and the latter to the third category described in \Cref{tab:models_cat}.
By combining \emph{n}-best lists from the L2R model, specialized at producing accurate prefixes with diverse suffixes, and from the R2L model, specialized at generating accurate suffixes with diverse prefixes, we aim to generate highly accurate but diverse \emph{n}-best lists.
\Cref{sec:pilot_generation} provides more details about our exploration.
For fair comparison, we employs the same L2R models as our baseline sequence-level KD experiments.

\subsection{Scaling Up \emph{n}-best Reranking} 

Deploying the complete set of scoring models to showcase accuracy improvements on a small set of test data is relatively affordable. 
However, deploying the same complete set to distill the entire training dataset becomes computationally intractable given the scale of the data.
Therefore, we describe two efforts to scale up \emph{n}-best reranking, namely reducing the number of scoring models at distillation time and reducing the number of student model's training data to distill.

\subsubsection{Model Reduction}
\label{sec:modelselection}

We seek to find a subset of scoring models, namely $\boldsymbol{\mathcal{D}}(\boldsymbol{s}, \boldsymbol{t}) \subset \boldsymbol{\mathcal{M}}(\boldsymbol{s}, \boldsymbol{t})$, that provides minimal quality degradation.
In our case, we think the goal is attainable since, despite the intended complementarity of the models, there may be significant redundancy, particularly as the majority of our in-house models are trained on the same data.

Manual selection of $\boldsymbol{\mathcal{D}}(\boldsymbol{s}, \boldsymbol{t})$ is impractical given the vast number of choices. 
Instead, we adopt a simple solution by leveraging the discriminatively learned weights $\boldsymbol{\lambda}$ associated with each scoring model. 
This approach capitalizes on the regularization term employed by the KB-MIRA optimizer (\Cref{sec:optimization}), offering a convenient and inexpensive way of selecting models that contribute significantly to the task. 
In our experiments in \Cref{sec:intrinsic}, we select top 5 models with the heighest weights for distillation, reducing the model count in our reranker with minimal accuracy drop.

\subsubsection{Transfer Set Reduction}
\label{sec:transferreduction}

The distillation cost of our \emph{n}-best reranker is proportional to the size of the so-called \emph{transfer set}, refers to the examples that were distilled and used to train the student model \cite{hinton-distilling}.
Typically, to maximize accuracy, the transfer set for NMT includes a new set of monolingual data, as suggested by \cite{edunov-etal-2018-understanding}.
This, in addition to the whole parallel data used also to train the teacher model, significantly increases the distillation cost.

To reduce the distillation cost, thus, we investigate transfer set reduction.
Particularly, in our experiments in \Cref{sec:reranker_improves_student}, we explore using distilled bitext, monolingual data or the combination of both. 
We found that using only the monolingual data as the transfer set is adequate with no accuracy drop, which leads to a significant saving in distillation time.
In addition, we also experiment with a significantly smaller transfer set that consists of the aggregate of multiple in-domain validation sets, used for finetuning a baseline student model, similar to \cite{finkelstein2023mbr}.
Unfortunately, our initial investigation suggests that while it does improve the baseline model's accuracy, the gain is marginal.

\section{Experimental Results}
\label{sec:experiments}

To showcase the efficacy of our proposal, we conduct large-scale experiments on WMT21 German $\leftrightarrow$ English and Chinese $\leftrightarrow$ English translation tasks. 
Our baseline is the vanilla sequence-level KD \cite{kim-rush-2016-sequence} that employs the aforementioned LR models as its teachers. 
We constrained the student model's capacity to approximately 68 million parameters, in line with the Transformer \emph{Base} architecture. 
We use Fairseq \cite{ott2019fairseq} for training and inference of our in-house models. More details about these models can be found in \Cref{sec:exp_setup}, including other experimental setup including the bitext used mainly for teacher model training and the monolingual data primarily used for student model training.  
We use the WMT19 set to learn $\boldsymbol{\lambda}$ weights for our reranker, the WMT20 set as our validation set and the WMT21 set as our blind test set. For these sets, we use the maximum number of references provided.
To report accuracy, we use sacreBLEU \cite{post-2018-call} with the following signature \texttt{nrefs:k|case:mixed|eff:no|tok:13a|smooth\\:exp} where \texttt{k} is the number of reference(s). 
For our main results, we additionally report chrF \cite{popovic-2015-chrf} with this signature \texttt{nrefs:k|case:mixed|eff:yes|nc:6|nw:0|spa\\ce:no} and COMET22 \cite{rei-etal-2022-comet} using \texttt{wmt22-comet-da} model. 
For generating \emph{n}-best list and student model's hypothesis, we set beam size to 8 and 5 respectively. 

In \Cref{sec:intrinsic}, we first focus on \emph{intrinsic} evaluation, comparing the accuracy of the \emph{n}-best reranker with that of the sequence-level KD's teacher models on validation sets. 
In \Cref{sec:reranker_improves_student} and \Cref{sec:self_training_improves_student}, we then shift to \emph{primary} evaluations where we assess the utility of the pseudo-labels generated by our \emph{n}-best reranker for training student model and retraining teacher models. 
We mainly focus on the German $\rightarrow$ English direction and summarize the results for the other language pairs at the end.

\subsection{Accuracy of \emph{n}-best Reranker}
\label{sec:intrinsic}
 
Table~\ref{tab:nbest_accuracy} summarizes the accuracy of our \emph{n}-best reranker on the German $\rightarrow$ English's validation set. 
In the WMT20 set, the baseline system attains a BLEU score of 58.8. 
This score also represents the score of the top-1 hypothesis in our \emph{n}-best list since the list is generated by the same model (complemented with its right-to-left counterpart).

In rows \emph{Oracle} and \emph{Anti-Oracle}, we report the accuracy of the best-scoring and worst-scoring hypotheses within our \emph{n}-best list.
Row Oracle shows that the best-scoring hypotheses surpass the top-1 by almost 10 BLEU point, indicating the substantial room for improvement embedded in our \emph{n}-best reranking approach.
Conversely, row Anti-Oracle shows that the gap to the worst-scoring hypotheses is much wider, which is almost 20 BLEU point worse.
This underscores the importance of employing robust scoring models, given the risk associated with poor-scoring alternatives. 

\begin{table}[ht]
    \centering
    \begin{tabular}{lr@{.}l}
    \toprule
        Description &  \multicolumn{2}{c}{\!\!\!\!\!WMT20} \\
    \midrule
         %0 & \multicolumn{3}{c}{$\boldsymbol{\mathcal{M}}$: All 72 models} & 60.4 \\ \cline{2-4}
          Baseline / Top-1 & 58 & 8 \\
         \hline
          Oracle & 67 & 5 \dag \\ 
          Anti-Oracle & 41 & 3  \\
          \hline         
          \emph{n}-best Reranker - Full ($|\boldsymbol{\mathcal{M}}|=72$) & 60 &4 \dag \\
          \emph{n}-best Reranker - Select ($|\boldsymbol{\mathcal{M}^d}|=5$) & 60 & 3 \dag \\
          \hline
          \hline
          \emph{k}NN-MT  & 59 & 1 \\
          MBR-BLEU  & 59 & 3 \\
          \bottomrule
    \end{tabular}
    \caption{\emph{N}-best reranker results on WMT20 validation. \dag implies that the difference is statistically significant with the Baseline at $p<0.05$.}
    \label{tab:nbest_accuracy}
\end{table}

Using the full set of 72 models ($\boldsymbol{\mathcal{M}}$), our \emph{n}-best reranker achieves the BLEU score of 60.4, surpassing the baseline system by 1.6 BLEU point.
This outcome underscores the efficacy of our \emph{n}-best reranker proposal in enhancing model accuracy. 
We then proceed to apply the model selection strategy described in \Cref{sec:modelselection}.
We pick 5 models with the highest weights, rerun reranking with the same weights (zeroing out the weights of other models) and report the reranker accuracy in the last row.
As shown, the accuracy of the \emph{n}-best reranker with smaller model count is relatively similar to running with the full set of models.

In the last rows of \Cref{tab:nbest_accuracy}, we also include two systems from \cite{yang-etal-2022-nearest} and \cite{finkelstein2024mbr} for reference.
The models from these two system are already in included in the scoring models in our \emph{n}-best reranker experiment.
The \emph{k}NN-MT is based on the vanilla \cite{khandelwal2021nearest} trained on the same training data as our baseline. 
For inference, we set $k=64$ and $\tau=100$. 
For MBR-BLEU, we use our baseline model to generate 260 hypotheses for each source sentences, where we use beam search to generate 4 hypotheses and use epsilon sampling with $\epsilon=0.02$ to generate the remaining 256 hypotheses, following \cite{finkelstein2024mbr}. 
As shown, the accuracy of these two systems are better than the baseline systems, but combining them with other models provide a much better accuracy. 
Considering the computational cost of \emph{k}NN-MT and MBR-based in generating hypotheses (discussed in \Cref{sec:distillation_cost}), we opt to include these methods as scoring models.

\begin{table*}[ht]
    \centering
    \begin{tabular}{cllccc}
    \toprule
        Rank & Description & Model Category & $\boldsymbol{\mathcal{G}}$ &  $\boldsymbol{\mathcal{D}}$ & WMT20 \\
        \midrule
         1 & FAIR WMT21 Dense & (8) Public model  & - & \checkmark & 59.6 \\
         %2 & MBR $\mathcal{U}$=BLEU  & (7) MBR loss & - & - & 59.0  \\ 
         %3 & MBR $\mathcal{U}$=chrF  & (7) MBR loss & - & - & 58.9  \\
         %4 & TransformerBig w/ MEGA & (1) Forward TM & - & - & 58.9 \\
         5 & TransformerBig L2R  & (1) Forward TM & \checkmark &- & 58.8 \\
         13 & TransformerBig R2L & (3) Right-to-Left TM & \checkmark & - & 58.0 \\
         14 & TransformerBig $d$=cc & (4) Adapted  & - & \checkmark & 58.0  \\
         19 & BigScience mt0-xxl-mt & (8) Public model & - & \checkmark& 57.8  \\         
         32 & TransformerBigBwd, R2L $d$=rapid & (2,3,4) Backward, R2L, Adapted & - & \checkmark& 54.8 \\
         50 & TransformerBigBwd, L2R & (2) Backward TM & - & \checkmark & 54.7 \\
         \bottomrule
    \end{tabular}
    \caption{Description of the models used to generate \emph{n}-best lists ($\boldsymbol{\mathcal{G}}$) and models selected for distillation ($\boldsymbol{\mathcal{D}}$), specifically for the first iteration of German $\rightarrow$ English direction, together with their accuracy on WMT20. }
    \label{tab:models_cat_acc}
\end{table*}

\begin{table*}[ht]
    \centering
    \begin{tabular}{clcccc}
    \toprule
         Row & Transfer Sets & Baseline & Seq-level KI & Seq-level KD & \emph{n}-best rerank \\         
         \midrule
         1 & bitext only (91M) & 48.8 & 49.3 & 49.6 & 50.0 \\
         2 & bitext + mono (155M) & - & - & 50.9 & 52.0 \\
         3 & mono only (54M) & - & - & 50.9 & 52.2 \\
         \bottomrule
    \end{tabular}
    
    \caption{Comparison of BLEU scores on WMT21 test sets between \emph{n}-best reranking and the three baseline models, including sequence-level knowledge interpolation and distillation, across different configurations of transfer sets. }
    \label{tab:student_only}    
    \vspace{-10pt}
\end{table*}
 
\Cref{tab:models_cat_acc} compiles the WMT20 accuracy of some models that are eventually utilized in the distillation of the student model's training data.
We rank the models based on the accuracy of each model when it is used as the \emph{only} model to rerank the \emph{n}-best list.
As shown, the two models utilized for generating the \emph{n}-best list ($\boldsymbol{\mathcal{G}}$) are ranked 5 and 13 respectively, but are not selected by our model selection strategy.
Interestingly, the models selected for distillation ($\boldsymbol{\mathcal{D}}$) exhibit considerable variability in terms of ranking, notably excluding the top highest-ranked models.
We hypothesize that this is due to redundancy in high-performing models, and the reranker prioritizes model diversity as also suggested by \cite{gontijo-lopes2022no}. 
The first model in $\boldsymbol{\mathcal{D}}$ is the single multilingual dense model provided by \cite{tran-etal-2021-facebook}, which is the most accurate model. 
While this model is not their final submission to WMT, it is highly accurate since it is trained on significantly larger training data and consists of 4.7 billion parameters. 
The remaining four other models in $\boldsymbol{\mathcal{D}}$ come from different model categories, ranging from backward, adapted, R2L and publicly-available models. 
Note that since the model selection strategy is automatic and non-deterministic, the models chosen for each iteration are dynamic. 
This is also applicable in other translation pairs. 

\begin{table*}[]
    \centering
    \begin{tabular}{lc| lcc | lcc}
        \toprule
         \multirow{2}{*}{System} & \multirow{2}{*}{$\mid \theta \mid$} & \multicolumn{3}{c|}{ German $\rightarrow$ English} & \multicolumn{3}{c}{ English $\rightarrow$ German} \\
         & & BLEU & chrF & COMET22 & BLEU & chrF & COMET22 \\
         \midrule
         1. Baseline & 68M & 48.8 & 67.4 & 84.8 & 52.6 & 68.3 & 82.0 \\
         2. Seq-level KD  & 68M & 50.9\dag & 68.6 & 85.7 & 54.5\dag & 69.1 & 83.0 \\
         \rowcolor{gray!25}
         3. \emph{n}-best (iter 1) & 68M & 52.2\dag\ddag & 69.3 & 85.9 & 57.4\dag\ddag & 71.1 & 84.4\\
         \rowcolor{gray!25}
         4. \emph{n}-best (iter 2) & 68M & 52.7\dag\ddag & 69.9 & 85.8 & 58.3\dag\ddag & 71.6 & 84.6 \\
         \rowcolor{gray!25}
         5. \emph{n}-best (iter 3) & 68M & 52.8\dag\ddag & 70.0 & 86.0 & 59.1\dag\ddag & 72.0 & 84.9 \\
         6. FAIR WMT21 Dense & 4.7B & 52.6\dag\ddag & 69.6 & 86.3 & 59.9\dag\ddag & 72.7 & 85.5 \\
         7. FAIR WMT21 MoE & 52B & 53.3\dag\ddag & 70.6 & 86.5  & 62.0\dag\ddag & 73.5 & 85.8 \\
         \bottomrule
    \end{tabular}
    \vspace{8pt}
    \begin{tabular}{lc| lcc | lcc}
        \toprule
         \multirow{2}{*}{System} & \multirow{2}{*}{$\mid \theta \mid$} & \multicolumn{3}{c|}{ Chinese $\rightarrow$ English} & \multicolumn{3}{c}{ English $\rightarrow$ Chinese} \\
         & & BLEU & chrF & COMET22 & BLEU & chrF & COMET22 \\
         \hline
         1. Baseline & 68M & 21.2 & 48.7 & 60.5 & 40.3 & 29.7 & 79.9 \\
         2. Seq-level KD  & 68M & 22.9\dag & 53.1 & 73.5 & 42.5\dag & 33.1 & 81.6 \\
         \rowcolor{gray!25}
         3. \emph{n}-best (iter 1) & 68M & 28.3\dag\ddag & 57.4 & 79.5 & 43.7\dag\ddag & 33.3 & 81.7\\
         \rowcolor{gray!25}
         4. \emph{n}-best (iter 2) & 68M & 29.4\dag\ddag & 58.1 & 80.2 & 45.2\dag\ddag & 34.9 & 82.7\\
         \rowcolor{gray!25}
         5. \emph{n}-best (iter 3) & 68M & 30.3\dag\ddag & 59.4 & 80.8 & 45.5\dag\ddag & 35.2 & 83.0\\
         6. FAIR WMT21 Dense & 4.7B & 29.9\dag\ddag & 60.1 & 81.9 & 42.4\dag\ddag & 34.3 & 85.2 \\
         7. FAIR WMT21 MoE & 52B & 32.1\dag\ddag & 60.4 & 82.2 & 49.9\dag\ddag & 39.4 & 85.2  \\
         \bottomrule
    \end{tabular}
    
    \caption{German $\leftrightarrow$ English (top) and Chinese $\leftrightarrow$ English (bottom) results on WMT21 test set, compared to the baseline models and WMT21 models from FAIR. FAIR MoE accuracy is from \cite{wmt21scores}. Our \emph{n}-best reranking results are in gray. $\dag$ implies that the difference with the baseline is statistically significant at $p < 0.05$, while $\ddag$ implies that the difference with the Seq-level KD is statistically significant at $p < 0.05$.}
    \label{tab:deende}
\end{table*}

\subsection{\emph{N}-best Reranking Improves Student}
\label{sec:reranker_improves_student}

This section investigates the utility of our \emph{n}-best reranking approach on the downstream task of training student model. 
We use the reranker with selected models ($\boldsymbol{\mathcal{D}}$) to generate the pseudo-labels for the whole training data. 
For our baseline sequence-level KD, we use the L2R model to generate the pseudo-labels. 
As another baseline, we also include sequence-level Knowledge Interpolation (KI) from \cite{kim-rush-2016-sequence}, which chooses hypotheses in the \emph{n}-best list that give the highest BLEU score using the original labels as the references. 

As part of scaling up mentioned in \ref{sec:transferreduction}, we explore how the accuracy of the student model is impacted by different \emph{transfer sets}. 
We investigate three configurations, namely \emph{bitext} only, \emph{bitext + monolingual}, and \emph{monolingual} only. 
Table~\ref{tab:student_only} summarizes the results of our experiments, which contains the accuracy of various student models on WMT21 test set. 

In the \emph{bitext only} condition, we only consider the \emph{distilled} parallel data to train the student model. 
More specifically, we compare the pseudo-labels generated by \emph{n}-best reranking with three baseline methods: original labels, pseudo-labels obtained through sequence-level knowledge interpolation (KI), and those obtained through sequence-level knowledge distillation (KD). 
As shown in row 1, the student model trained with the original labels achieved an accuracy of 48.8 BLEU point. 
Meanwhile, the models trained with pseudo-labels generated through sequence-level KI and KD showed improvements of 0.5 and 0.8 BLEU points respectively, which is in line with previous literature \cite{kim-rush-2016-sequence}. 
Our \emph{n}-best reranker approach leads to even stronger performance, with the student model achieving an accuracy of 50.0 BLEU point. 
This is a statistically significant improvement of 1.2 BLEU points compared to the baseline.

In the \emph{bitext+mono} condition, we augment the training data for the student model with the {distilled} monolingual data. 
Since the monolingual data lack labels, we compare our \emph{n}-best reranking method only with sequence-level knowledge distillation (KD). 
The results in row 2 reveal that incorporating the distilled monolingual data significantly improves the accuracy of the sequence-level KD system by approximately 1.3 BLEU points. 
However, our \emph{n}-best reranking approach achieves an even greater gain of 2.0 BLEU points, thereby widening the performance gap with sequence-level KD to 1.1 BLEU points. 
This result highlights the value of incorporating in-domain data as the transfer sets. 
In our approach, the monolingual data used seems to align with the domain of the evaluation sets, while in contrast, the parallel data are sourced from a broader range of domains. 

In row 3, we investigate whether a smaller in-domain transfer set is more or as effective than a larger mixed-domain one. 
The results in row 3 reveal marginal gains for both sequence-level KD and our \emph{n}-best reranking approach when using only the distilled monolingual data as the transfer sets. 
This result is highly encouraging since we can reduce the distillation time by half without accuracy drop.
In any case, our \emph{n}-best reranking approach leads to a student model that is 3.4 BLEU better than the baseline and 1.3 BLEU better than sequence-level KD.
Based on these results, we use monolingual data as the transfer set subsequently.

\subsection{Self-Training Teacher Improves Student }
\label{sec:self_training_improves_student}

Given the substantial accuracy gains obtained by using pseudo-labels generated by \emph{n}-best reranker in student models, we investigate whether teacher models can derive similar benefit from the use of the same pseudo-labels. 
Up to now, all the teachers models are trained exclusively from parallel data with original labels that come from a mixed set of domains. 
In light of this, we conduct a series of experiments retraining the teacher model using pseudo-labels. 
To manage computational costs effectively, we focus our investigations on retraining the models in $\boldsymbol{\mathcal{G}}$.
This is accomplished through fine-tuning the models, as opposed to retraining them from scratch, and utilizing only monolingual data, excluding the bitext, as the transfer sets. 
Our rationale for this strategy is detailed in the preliminary experiments, discussed in the \Cref{sec:finetuning_or_retraining}.

More specifically, we \emph{fine-tune} the two models in $\boldsymbol{\mathcal{G}}$ for one epoch using the pseudo-labels obtained from the \emph{n}-best reranker and using monolingual data as the transfer sets. 
We then retrain the next iteration's reranker using these models, producing a new set of pseudo-labels for training the student model. 
It's worth reiterating that the models selected for distillation $\boldsymbol{\mathcal{D}}$ vary in each iteration. 
We continue this iterative process \emph{twice} when we typically start observing diminishing gain. 

Table~\ref{tab:deende} provides a summary of our self-training experiments. 
Focusing on the German $\rightarrow$ English columns, the first three rows of the table are taken from Table~\ref{tab:student_only}, reporting the accuracies of the baseline model, the student model trained with sequence-level KD, and the student model trained with pseudo-labels from \emph{n}-best reranking. 
The next two rows show the results from our self-training experiments for two iterations. 
Our experiments show that self-training the teacher models for one iteration can improve the student model accuracy by 0.5 BLEU points (row 4). 
Our final model, after three iterations, scores 4.0 BLEU points higher than the baseline model and 2.9 BLEU points higher than sequence-level KD. 
This conclusion is consistent across both chrF and COMET metrics.
We also compare our final model with the winning WMT21 models from FAIR with respect to accuracy and model size, as shown in rows 6 and 7. 
Performance-wise, our final model is comparable to FAIR’s Dense model, while having fewer parameters. Our model consists of 68 million parameters, while the FAIR model is around 70 times larger. 

We also present the experimental results for the English $\rightarrow$ German and the Chinese $\leftrightarrow$ English directions in \Cref{tab:deende}. 
As shown, we observe a gain similar to the one observed in the German $\rightarrow$ English direction where the pseudo-labels from \emph{n}-best reranker leads to a significantly better student accuracy.
These gains remain consistent across multiple metrics, encompassing chrF and COMET22, although given that our reranker is trained to optimize BLEU score, the most pronounced improvement is evident in the BLEU score.
Nevertheless, these results affirm our hypothesis that the \emph{n}-best reranker with robust scoring models can effectively enhance the quality of training data labels. 

\section{Related Work}

Our proposal intersects with many works in various ways. The idea of utilizing \emph{n}-best reranking to improve accuracy has been extensively investigated as far back as the era of Statistical Machine Translation if not earlier, for example in \cite{och-etal-2004-smorgasbord,shen-etal-2004-discriminative,chiang-etal-2008-online} and more recently in \cite{marie-etal-2020-combination,qian-etal-2021-volctrans,tran-etal-2021-facebook}. 
In these recent work, \emph{n}-best reranking incurs significantly higher inference time from running multiple models over the \emph{n}-best list, thus may not be practical for real-world systems. 
In contrast, our work makes a practical trade-off by shifting the heavy computational cost of \emph{n}-best reranking to training data preprocessing without affecting the latency of the deployed model. 
Our work shares the same motivation as \cite{yang-etal-2022-nearest,finkelstein2023mbr}, but we consider a larger and more diverse set of models. % to include our \emph{n}-best list. 

The idea of looking at \emph{n}-best hypotheses for knowledge distillation has been also investigated in the original sequence-level KD paper \cite{kim-etal-2021-distilling-knowledge}, namely sequence-level Knowledge Interpolation which we consider as one of our baseline where the authors propose to approximate the mode with the hypothesis that scores the highest according some translation metrics. 
However, since this approach requires the ground truth, the application of this variant is limited to distilling parallel data. 
In contrast, since our \emph{n}-best reranker is trained on a tune set, our approach is applicable for distilling unlabelled monolingual data.

Our \emph{n}-best reranker incorporates various models as reranking models. 
Some of these models have been applied to knowledge distillation. 
For example, \newcite{yang-etal-2022-nearest} deploys nearest neighbor machine translation models. Meanwhile, \newcite{yee-etal-2019-simple} combines direct models with channel and language models. 
\newcite{currey-etal-2020-distilling} trains domain-specific teacher models to distilled in-domain training data for training multi-domain student model. 
On the other hand, our \emph{n}-best reranker incorporates significantly larger number of models, including the aforementioned. 
Also, we deploy these models to score hypotheses, rather than to generate them, which is significantly faster.

Self-training has also been frequently investigated for Machine Translation in statistical and neural era \cite{li-etal-2019-niutrans}. 
Recently, it is often dubbed as iterative knowledge distillation and can be found as a winning formula in many evaluation campaigns \cite{li-etal-2019-niutrans}. 
In this work, we apply self-training using high-quality pseudo-labels from \emph{n}-best reranker which produces accurate results. 

\section{Summary and Future Work}

We enhance the sequence-level knowledge distillation \cite{kim-rush-2016-sequence} by incorporating \emph{n}-best reranking. 
Thus, rather than improving the accuracy of the teacher models by following neural scaling laws alone, our proposed method do so by leveraging a multitude of models with different inductive biases, objective functions or architectures to collaboratively rescore \emph{n}-best hypotheses and identify the best pseudo-labels. 
Furthermore, we observed a relatively strong cascading effect, where teacher models finetuned using pseudo-labeled data are more accurate, leading to the generation of more accurate pseudo-labels for the next iteration and resulting in an even more accurate student model. 
We also explore efficienty efforts to scale up \emph{n}-best reranking via model selections and transfer set reductions, resulting in a reduction in distillation time. 
Our final student model demonstrates up to 4.0 BLEU point improvement over baseline systems and is on par with a strong large translation model on German$\leftrightarrow$English and Chinese$\leftrightarrow$English translation tasks, despite having only 1/\nth{70} the parameters.

For future work, we intend to improve the efficacy of our approach by incorporating more powerful large language models that are finetuned towards translation tasks as well as models that more explicitly capture fine-grained phenomena such as gender or number agreements.
To improve the efficiency, we also plan to investigate methods to automatically identify transfer sets at fine-grained sentence level as well as ways to speed up the scoring process further, for instance by utilizing only unnormalized probabilty score like in \cite{devlin-etal-2014-fast}.

\section*{Limitations}
While the proposed evaluation framework is language-agnostic, the experiments conducted in this study are limited to two language pairs. Due to its reliance on the availability of models and in-domain monolingual, we cannot guarantee accurate results when applied to language pairs involving a low-resource language pairs.
We use numerous pretrained models with various license terms. While all of them are friendly for non-commercial research purpose, not all of them are not for commercial purpose. Readers should perform their own due dilligence. 

\section*{Ethics Statement}

We acknowledge the ethical considerations associated with the \emph{n}-best reranking approach, which utilizes multiple models to generate pseudo-labels. First of all, we recognize that these models possess their own biases, inherited from the training data, which can potentially perpetuate societal inequalities. Bias in the models can result from biased training data or the inherent limitations of the algorithms used. Despite our best efforts to preprocess and debias the training data, complete elimination of biases is challenging. Second of all, the \emph{n}-best reranking approach incurs higher computational costs compared to traditional methods. These costs arise from training and maintaining multiple models concurrently. We have implemented mitigation strategies such as model recycling and leveraging publicly available corpora to address these concerns. Furthermore, although we utilize numerous models, it is important to highlight that the majority were not specifically trained for this \emph{n}-best approach. In fact, many originate from our broader exploratory initiatives, and the \emph{n}-best reranker serves as a means to repurpose them effectively. Despite of our mitigation efforts, the increased computational burden can limit the accessibility and affordability of the approach, particularly for researchers or organizations with limited resources.

\bibliography{anthology,custom}

\appendix
\section{Experimental Setups}
\label{sec:exp_setup}

We follow the experimental setup of the WMT21 news translation task, particularly the constrained track to train our in-house models. 
For German $\leftrightarrow$ English directions, our parallel data are composed of Europarl v10, ParaCrawl v7.1, Common Crawl, News Commentary v16, Wiki Titles v3, Tilde Rapid and WikiMatrix. 
For Chinese $\leftrightarrow$ English directions, our parallel data are composed of Paracrawl v7.1, News Commentary v16, Wiki Titles v3, UN Parallel Corpus v1.0, CCMT and WikiMatrix.
For monolingual data, we use the 2021 subsets of News Crawl.  We deduplicate and preprocess the data using the M2M-100 (Fan et al., 2021) processing scripts\footnote{\url{https://github.com/facebookresearch/fairseq/tree/main/examples/m2m_100}}.
For training our in-house teacher models, we run up to 80 thousand updates, while for training the student model, we run up to 30 thousand updates. For finetuning teacher models, we run one epoch of updates.

\Cref{tab:data_statistics} summarizes the data sizes for different splits of the two language pairs.
\begin{table}[h]
\begin{tabular}{crrrr}
\toprule
    & Bitext & Mono & Valid & Test \\
    \midrule
     De $\rightarrow$ En & 91M & 38M & 785 & 1000 \\
     En $\rightarrow$ De & 91M & 37M & 1418 & 1002 \\
     Zh $\rightarrow$ En & 54M & 32M & 2000 & 1948 \\
     En $\rightarrow$ Zh & 54M & 37M & 1418 & 1002 \\     
     \bottomrule
\end{tabular}
\caption{Sizes of Splits used in the paper, where Valid refers to WMT20 and Test refers to WMT21.}
\label{tab:data_statistics}
\end{table}

\section{Pilot Study for Models for \emph{n}-best Generation}
\label{sec:pilot_generation}

Figure~\ref{f:beam} shows the BLEU scores for the top-1 and oracle hypotheses of \emph{n}-best list with different \emph{N} from 1 to 32 on our tune set. As shown, the BLEU score for the top-1 hypotheses marginally improves when we increase the beam size from 1 to 4 but then it plateaus, which is consistent with \newcite{britz-etal-2017-massive}'s finding. This suggests that increasing the beam size may not benefit the original sequence-level KD. In contrast, the oracle BLEU score improves monotonically with larger beam size, where the gap for $N>8$ is more than 10 BLEU points and growing. This gap speaks to the potential for our proposed \emph{n}-best reranking. Compared to L2R, the \emph{n}-best list's oracle score from L2R+R2L is around 2-3 BLEU points higher. We equate $\boldsymbol{\mathcal{G}}$ to L2R+R2L with beam size of 8 since its accuracy is better than doubling the beam size of L2R setup with additional parallelization benefits. 

\pgfplotsset{compat = newest}
\pgfplotstableread{beam.dat}{\beamtable}
\begin{figure}
\centering
\begin{tikzpicture}
\begin{axis}[
     xmin=0,
    major grid style = {lightgray},
    minor grid style = {lightgray!25},
    width = 0.5 * \textwidth,
    xtick={1,2,4,8,16,32},
    xlabel={beam size},
    ylabel={BLEU},
    legend style={at={(0.35,0.2)},anchor=south west}   
]
 
\addplot[teal, mark = x, mark size = 3pt] table [x = {beam}, y = {top1}] {\beamtable};
\addplot[teal, mark = o, mark size = 3pt] table [x = {beam}, y = {L2R-oracle}] {\beamtable};
\addplot[teal, mark = +, mark size = 3pt] table [x = {beam}, y = {L2R+R2L-oracle}] {\beamtable};
\legend{
    top1, 
    L2R (oracle),
    L2R+R2L (oracle)
}
 
\end{axis}
 
\end{tikzpicture}
\caption{BLEU scores for top-1 and oracle hypotheses of WMT19 with different beam size} \label{f:beam}
\end{figure}
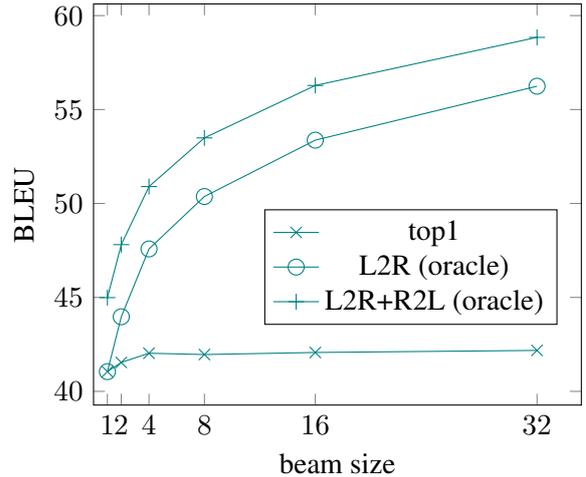

\section{Pseudo-Labels from \emph{N}-best reranking for Self-Training}

\label{sec:finetuning_or_retraining}

We conduct a pilot study on \emph{one} of the L2R models, which is part of $\boldsymbol{\mathcal{G}}$, to inform our decisions on two aspects: 1) determining which transfer sets to utilize, and 2) determining whether it is necessary to retrain the teacher model from the scratch or if fine-tuning proves to be sufficient.
The results of this pilot study are summarized in Table~\ref{tab:teacher_pseudo}. 
For finetuning, we only run one epoch, while for retraining we run around 50 epochs (up to 80 thousands update).
The baseline accuracy of training this teacher model using the original bitext is 57.4 point, as indicated in the first row of column \emph{Baseline}. 
The \emph{Retrain} column shows that training the teacher model with the same bitext, but with pseudo-labels, resulted in a gain of 0.6 BLEU point. 
As shown in the subsequent rows (bitext+mono and mono only), adding the distilled monolingual data to the transfer sets or using them alone result in a stronger gain of around 1.5 BLEU points, which is consistent with our finding in the student model training.

\begin{table}[h]
    \centering
    \begin{tabular}{lccc}
        \toprule
          Transfer Sets & Baseline & Retrain & Finetune  \\
         \midrule
         bitext only & 57.4 & 58.0 & 58.0 \\
         bitext+mono & - & 59.5 & 59.2 \\
         mono only & - & 59.5 & 59.5 \\
         \bottomrule
    \end{tabular}
    \caption{WMT20 scores of a teacher model trained with pseudo-labels from \emph{n}-best reranking with different transfer sets (rows) and training regime (columns).}
    \label{tab:teacher_pseudo}
\end{table}

Comparing the Retrain and Finetune columns, we observe that the accuracy of finetuned models is on par with the model trained from scratch. These results are encouraging because we can obtain a teacher model that is 2.1 BLEU points more accurate with minimal training FLOPs via finetuning and using the smallest transfer set. We conduct similar experiments using pseudo-labels from sequence-level KD and discuss it in Appendix~\ref{sec:appendix:seq_self}. Although a similar trend is observed, the resulting gain from sequence-level KD is smaller.

\section{Pseudo-Labels from Sequence-level KD for Self-Training}
\label{sec:appendix:seq_self}

We report the results for self-training teacher model using the pseudo-labels from sequence-level KD in Table~\ref{tab:teacher_pseudo_seq}. As shown in row \emph{bitext only}, retraining teacher models with these pseudo-labels leads to a degradation. Including the monolingual data as the transfer sets helps to improve the accuracy as shown in row bitext+mono and mono only. Comparing columns Retrain and Finetune, we observe that finetuning can achieve a similar accuracy gain as the full retraining, which is similar to what we observe in finetuning experiments using pseudo-labels from \emph{n}-best reranking. Comparing with using pseudo-labels from \emph{n}-best reranking reported in Table~\ref{tab:teacher_pseudo}, self-training using pseudo-labels from sequence-level KD gives smaller accuracy gain than self-training using pseudo-labels from \emph{n}-best reranking. 

\begin{table}[h]
    \centering
    \begin{tabular}{lccc}
    \toprule
          & Baseline & Retrain & Finetune  \\
         \midrule
         bitext only & 57.4 & 57.3 & 57.1 \\
         bitext+mono & - & 58.3 & 57.8 \\
         mono only & - & 58.3 & 58.1 \\
         \bottomrule
    \end{tabular}
    \caption{WMT20 BLEU scores of a teacher model trained with pseudo-labels from \emph{sequence-level KD} with different transfer sets (rows) and training regime (columns).}
    \label{tab:teacher_pseudo_seq}
\end{table}

\section{Effects of Pseudo-Labels on Different Model Architectures}

This section describes the efficacy of pseudo-labels generated by \emph{n}-best reranker on the teacher models, beyond the student model described in the main paper. 
\Cref{tab:teacher_student_iterations} details the accuracy of teacher models with and without reranking along with the accuracy of the corresponding student models, focusing on the German $\rightarrow$ English WMT21 test set.
For the student models, we copy the numbers from the German $\rightarrow$ English section of \Cref{tab:deende}.
At iteration 0, the student model is trained with the original labels of the bitexts, while at the later iterations, the student is trained with monolingual data with pseudo-labels generated using \emph{n}-best reranker at the corresponding iteration. 
For column Top-1, we report the accuracy of the generating models, which refer to an ensemble of 4 models with Transformer Big architecture, each consisting of around 310 million parameters.
At iteration 1, these teacher models are trained with original labels of the bitexts, and at later iterations, they are trained with monolingual data with pseudo labels generated by the reranker at the previous iteration.

As shown, there is a 2 BLEU point gap between the student model (row iter 1; column Student) and the teacher models (row iter 1; column Top-1) when the two models are trained with parallel data with original labels.
As shown, the \emph{n}-best reranker improves the teacher accuracy by +2.5 BLEU point (from 50.8 to 53.3).
When this reranker is used to generate the pseudo-labels of the monolingual data for student model training, the student model's accuracy increases up to 52.2 BLEU score (row iter 2; column student).
When the same pseudo-labels are used to train the teacher models, the accuracy of teacher model's next iteration increases up to 52.5 BLEU point (row iter2 and column Top-1). 
A similar trend but with less significant improvement is also observed for iteration 2 and iteration 3.
This result demonstrates that the accuracy gain observed in the student model is also observed in the teacher models, which is an order magnitude larger. 
Additionally, it also shows that the accuracy gap between teacher and student models is smaller with the combination of self-training and \emph{n}-best reranking.

\begin{table}[h]
    \centering
    \begin{tabular}{cc|ccc}
    \toprule
         \multirow{2}{*}{Iter} &  \multirow{2}{*}{Student} & \multicolumn{3}{c}{Teacher}  \\
         \cline{3-5}
         & & Top-1 & Reranked & $\Delta$ \\
         \midrule
         1 & 48.8 & 50.8 & 53.3 & +2.5 \\
         2 & 52.2 & 52.5 & 53.6 & +1.1 \\
         3 & 52.7 & 53.0 & 54.0 & +1.0 \\
         4 & 52.8 & & & \\
         \bottomrule
    \end{tabular}
    \caption{WMT21 BLEU scores of teacher and student models across different iteration for the WMT21 German-English test set.}
    \label{tab:teacher_student_iterations}
\end{table}
 
\section{Distillation Cost}
\label{sec:distillation_cost}
\begin{table}[h]
    \centering
    \begin{tabular}{l@{\hspace{-3pt}}c@{\hspace{3pt}}c@{}}
        \toprule        
        \multirow{2}{*}{Step} & Parallel & Serial \\
        & Hours & Hours \\
        \toprule
        Generating \emph{n}-best & 00:48 & 01:33 \\
        \hline
        \, TransformerBig L2R & \multicolumn{2}{c}{00:48} \\
        \, TransformerBig R2L & \multicolumn{2}{c}{00:45} \\
        \midrule
        Scoring & 02:08 & 06:22\\
        \hline
         \, FAIR WMT21 Dense & \multicolumn{2}{c}{01:55} \\
         \, TransformerBig $d$=cc & \multicolumn{2}{c}{00:43} \\
         \, BigScience mt0-xxl-mt & \multicolumn{2}{c}{02:08}  \\         
         \, TransformerBigBwd,R2L,$d$=rapid & \multicolumn{2}{c}{00:51} \\
         \, TransformerBigBwd,L2R & \multicolumn{2}{c}{00:45} \\
         \midrule
         $\arg\max$ & \multicolumn{2}{c}{00:10*} \\
        \midrule
         \emph{n}-best reranking total & 03:06 & 08:05 \\
         \midrule
         \midrule
        \emph{k}NN-NMT & \multicolumn{2}{c}{14:54} \\
        MBR-BLEU &  \multicolumn{2}{c}{15:28} \\
         \bottomrule    
    \end{tabular}
    \caption{GPU hour breakdown of a German-English distillation process for a sample of 10,000 German sentences using \emph{n}-best reranker described in \Cref{tab:models_cat}. Last two rows report the distillation cost for two other methods. * refers to CPU hours}
    \label{tab:distillation_cost}
\end{table}

\Cref{tab:distillation_cost} presents the distillation cost incurred for distillating a sample of 10,000 German sentences utilizing the \emph{n}-best reranker outlined in \Cref{tab:models_cat}. Since our \emph{n}-best reranking consists of many parallelizable components, we detail the costs in terms of \emph{parallel} and \emph{serial} hours. Parallel hours depict a scenario in which all computing resources are accessible simultaneously, while serial hours depict a scenario where only one resource is available at a time. The actual wallclock time is contingent upon the condition of the compute cluster condition, which impacts the actual level of parallelism. Apart from the reranking step which computes the final cost of each hypothesis, all the cost in \Cref{tab:distillation_cost} refers to GPU hours. 

As shown, generating the \emph{n}-best list takes around one and a half hour to generate using the two generation models. 
For the baseline knowledge-distillation, utilizing solely the L2R model, the pseudo-labels can be generated in less than an hour. 
The majority of cost for \emph{n}-best stems from the scoring step, involving 5 models. 
The cost for each model correlates with the model size. Scoring using a 13 billion models (BigScience mt0-xxl-mt) takes around 2 hours while scoring using an ensemble of 4 models with 655 million parameters (TransformerBig) takes around 45 minutes. 
The parallel cost for the scoring step is around 2 hours, which represents an optimal situation where the computing resources are available to score using all models, which is equal to the time needed for the slowest model. 
Meanwhile, the serial cost for this step takes up eight hours, which represents a less than ideal situation where each model must wait for resources sequentially. 
The last step ($\arg\max$) is to takes the scored \emph{n}-best and rerank it according to the learned weights. 
This step only requires CPU and the cost is negligible compared to other steps.
In practice, the \emph{n}-best reranking takes between 3:06 to 08:05, incurring around 4.0x to 10x times distillation cost than the baseline sequence-level KD. 

In the last two rows, we report the distillation costs from two related work, namely \emph{k} Nearest Neighbor-(\emph{k}-NN NMT) from \cite{yang-etal-2022-nearest} and Minimum Bayes Risk decoding (MBR-BLEU) from \cite{finkelstein2024mbr}.
For \emph{k}NN-NMT, we generate the pseudo-labels with beam size of 8 similar with the baseline and set $k$ to 64 neighbors and the temperature $\tau$=100. 
For MBR-BLEU, following \cite{finkelstein2024mbr}, we generate 260 hypotheses for each source sentences by generating 4 hypotheses via beam search with beam size of 4 and the remaining 256 hypotheses via epsilon sampling with $\epsilon=0.02$. 
As shown, the distillation costs for these two related methods are substantially larger than our \emph{n}-best reranker approach.
The cost associated with \emph{k}NN-NMT is consistent with the conclusion of \cite{khandelwal2021nearest} where they reported two order magnitude slower inference speed than their baseline.
Meanwhile, the cost associated with MBR-BLEU is due to the high number hypotheses generated in the second decoding stage, which requires us to reduce the effective batch size significantly during inference time.

Lastly, while distilling with \emph{n}-best reranking introduces a notable increase in computational cost, it however offers a substantial improvement in accuracy. 
Thus, it is essential to acknowledge that while the cost of \emph{n}-best reranking may be higher, it pales in comparison to the labor-intensive process of manually creating new parallel data. 
Therefore, the expense incurred by \emph{n}-best reranking should be considered within the context of its significant accuracy gains and the resource-intensive alternative of generating new parallel data manually.

\end{document}